\documentclass[10pt,twocolumn,letterpaper]{article}

\usepackage[pagenumbers]{wacv} 

\usepackage{graphicx}
\usepackage{multirow}
\usepackage{wrapfig,lipsum}
\usepackage{enumitem}
\usepackage{dsfont}
\usepackage{adjustbox}

\newcommand{\cM}{\mathcal{M}}
\newcommand{\cD}{\mathcal{D}}
\newcommand{\cA}{\mathcal{A}}
\newcommand{\cE}{\mathcal{E}}

\newcommand{\cL}{\mathcal{L}}
\newcommand{\cY}{\mathcal{Y}}
\newcommand{\bx}{\mathbf{x}}

\newcommand{\pname}{NOVO}

\definecolor{wacvblue}{rgb}{0.21,0.49,0.74}
\usepackage[pagebackref,breaklinks,colorlinks,allcolors=wacvblue]{hyperref}

\def\wacvPaperID{*****} 
\def\confName{WACV}
\def\confYear{2026}

\title{{\pname}: Unlearning-Compliant Vision Transformers}

\author{Soumya Roy\thanks{Equal Contribution}\\
Amazon India\\
Bangalore, India\\
{\tt\small meetsoumyaroy@gmail.com}
\and
Soumya Banerjee$^*$\\
IIT Kanpur\\
Kanpur, India\\
{\tt\small soumyab@cse.iitk.ac.in}
\and
Vinay Verma$^*$\\
Amazon India\\
Bangalore, India\\
{\tt\small vinayugc@gmail.com}
\and
Soumik Dasgupta\\
Walmart Labs\\
Bangalore, India\\
{\tt\small soumikdasguptasl@gmail.com}
\and
Deepak Gupta\\
Amazon India\\
Bangalore, India\\
{\tt\small deepakgupta.cbs@gmail.com}
\and
Piyush Rai\\
IIT Kanpur\\
Kanpur, India\\
{\tt\small piyush@cse.iitk.ac.in}
}

\begin{document}
\maketitle
\begin{abstract}
Machine unlearning (MUL) refers to the problem of making a pre-trained model selectively forget some training instances or class(es) while retaining performance on the remaining dataset. Existing MUL research involves fine-tuning using a forget and/or retain set, making it expensive and/or impractical, and often causing performance degradation in the unlearned model. We introduce {\pname}, an unlearning-aware vision transformer-based architecture that can directly perform unlearning for future unlearning requests without any fine-tuning over the requested set. The proposed model is trained by simulating unlearning during the training process itself. It involves randomly separating class(es)/sub-class(es) present in each mini-batch into two disjoint sets: a proxy forget-set and a retain-set, and the model is optimized so that it is unable to predict the forget-set. Forgetting is achieved by withdrawing keys, making unlearning on-the-fly and avoiding performance degradation. The model is trained jointly with learnable keys and original weights, ensuring withholding a key irreversibly erases information, validated by membership inference attack scores. Extensive experiments on various datasets, architectures, and resolutions confirm {\pname}'s superiority over both fine-tuning-free and fine-tuning-based methods.
\end{abstract}
    
\section{Introduction}
\label{sec:introduction}

As AI agents become increasingly pervasive in our everyday lives, concerns about privacy, regulatory compliance, and legal requirements are growing, leading to requests for data removal. Stakeholders may also demand the removal of data instances of specific class(es) or categorie(s) from training dataset, eliminating its impact on the models~\cite{chundawat2023zero}. The laws like \textit{Right to be Forgotten}, GDPR~\cite{gdpr}, and CCPA~\cite{ccpa} further mandate such requirements. A na\"ive solution involves retraining from scratch, excluding those specific instances. While it yields optimal results, it is computationally expensive. With the rise of LLMs and LVMs~\cite{li2023blip,dong2024internlm,liu2024visual} like ChatGPT~\cite{achiam2023gpt}, LLAMA~\cite{touvron2023llama}, and CLAUDE~\cite{wu2023comparative}, repeated retraining is infeasible.

Currently, Machine Learning (ML) models like ViT~\cite{dosovitskiy2020image} and SWIN~\cite{liu2021swin} are trained for optimal performance without considering possible future unlearning requests. Recently, there has been a growing interest in MUL research~\cite{golatkar2021mixed,tarun2023fast,tarun2022deep,wu2020deltagrad,guo2020certified,graves2021amnesiac,lin2023erm,schelter2019amnesia,fan2024challenging,liu2024model,kurmanji2024towards,chen2023boundary,tarun2023deep,shen2024label,oesterling2024fair,kim2024layer,shao2024federated,han2024towards}, with the existing approaches mainly building on top of standard ML approaches, requiring fine-tuning with the instances for which unlearning has been requested. While this process enables unlearning, it often leads to reduced performance over the retained set, necessitating additional fine-tuning with the retain set, raising concerns about data privacy and storage. Moreover, it also incurs a high cost due to repeated fine-tuning. This prompts an important question: \emph{Is it possible to develop an unlearning-compliant model that can forget on demand over the requested classes or sub-classes without the need for explicit fine-tuning with data instances corresponding to the forget classes, while preserving the performance on the retain dataset/classes?}

This paper addresses the aforementioned question by proposing a novel architecture, {\pname} (U\textbf{\underline{N}}learning-C\textbf{\underline{O}}mpliant \textbf{\underline{V}}ision Transf\textbf{\underline{O}}rmers). Here, we focus on sub-class and full-class unlearning~\cite{lin2023erm,chundawat2023zero,tarun2023fast,yan2022arcane,liu2022continual,panda2024partially}, where instead of forgetting individual samples, the model forgets the entire sub-class(es) or class(es). For instance, in applications like face recognition, a face ID may represent a class, making class-level unlearning useful~\cite{chundawat2023zero}. Sub-class forgetting is useful when we want to forget entire sub-class(es) within a set of super-classes. The proposed model is simple and requires minor modifications to the standard architectures, which can be applied to any transformer-based architecture, such as ViT~\cite{dosovitskiy2020image}, CAIT~\cite{touvron2021going} and SWIN~\cite{liu2021swin}. The model is trained to be unlearning compliant and aware of potential future unlearning requests. \emph{It can forget any set of class(es) or sub-class(es) \emph{on-the-fly}, just by referencing the set of class/sub-class indices that have been requested to be forgotten. {\pname} does not require any forget set or retain set samples and is also free from any fine-tuning or expensive post-processing during the unlearning step.}

{\pname} associates a learnable key with each class or sub-class, enabling the model to predict classes only when the corresponding keys are present. The model has two sets of parameters: \emph{learnable keys and the original model weights, both of which are trained jointly} to form the final model. \emph{Therefore, the original weights alone do not have any prediction ability, which helps mitigate potential attacks and privacy concerns.} The removal of class keys results in permanently forgetting a set of classes or sub-classes. To achieve this, the model is trained by simulating learning and unlearning during the training process, where classes in each mini-batch are divided into proxy forget and retain sets, represented by two multi-hot binary vectors. During training, the model receives these vectors along with standard inputs, which helps select the corresponding class/sub-class keys. The composite parameters are optimized to minimize the cross-entropy loss over the retain set, while logits for the forget set are forced to follow a uniform distribution to simulate unlearning. By alternating between the retain and forget sets, this approach effectively simulates various unlearning scenarios. However, iterating over all possible combinations of retain and forget classes requires simulating $2^{C}$ combinations for $C$ classes, making it infeasible within a limited number of epochs. Therefore, high generalization ability is crucial to handle such unseen configurations. We empirically observe that our unlearning-aware approach to training {\pname} generalizes to unseen retain/forget set configurations, automatically returning a class near the vicinity of the forget set. Additionally, we observe that injecting the learnable keys into deeper layers of the network helps learn enriched class or sub-class-specific representations and enhances the model's performance. Furthermore, we introduce random dropping and expansion of the proxy retain and forget set multi-hot binary vectors to prevent any correlations between the learnable keys and the retained set samples, significantly boosting the model's robustness and overall accuracy.

{\pname} can also function as a controllable network, using class or subclass keys to control output predictions, making it suitable for scenarios requiring predictions only on a subset of classes or subclasses. Extensive experiments on multiple datasets involving single and multiple-class forgetting scenarios demonstrate that as the forget set grows, the performance of SOTA baselines degrades while our model remains unaffected. Our main contributions can be summarized as follows:
\begin{itemize}
    \item We propose a novel, on-the-fly, unlearning-aware transformer architecture that can forget the desired set of class(es) or sub-class(es) without requiring any expensive post-processing or training samples. 
    

    \item The deeper key injection, random drop and expand, and unlearning-aware training enhance learning and unlearning. {\pname} can work as a controllable network and supports predictions for a subset of class(es).
    

    \item Extensive experiments on CIFAR10/20/100, and TinyImagenet-200 using ViT~\cite{dosovitskiy2020image}, CAIT~\cite{touvron2021going}, and SWIN~\cite{liu2021swin} across resolutions (32$\times$32, 224$\times$224, 384$\times$384) show our model's efficacy over SOTA baselines, with ablation validating each component. The MIA score confirms robustness against inference attacks.
    
\end{itemize}
\section{Related Work}
\label{sec:related_work}

Early attempts to MUL~\cite{cao2015towards} introduced unlearning to eliminate the impact of specific data points on an already trained model. With the advent of stricter data privacy policies~\cite{ccpa,gdpr}, there has been a growing interest in MUL research. These approaches can be broadly classified as \textit{exact unlearning} and \textit{approximate unlearning} approaches.

\emph{Exact unlearning}~\cite{bourtoule2021machine,thudi2022unrolling} approaches remove the desired samples during retraining to achieve optimal performance on the retain and forget set. However, it is computationally expensive due to requiring repeated retraining. \emph{Approximate unlearning} addresses this by simulating a model trained without retraining it from scratch, using fewer updates. This is achieved by estimating the influence of forget-set samples and updating the model parameters accordingly. In this section, we focus on two key variants of approximate unlearning: \emph{zero-shot unlearning} and \emph{retraining-free unlearning}.

Current SOTA methods like~\cite{chundawat2022can,graves2021amnesiac, jia2024model} require access to at least a subset of the training dataset. However, in practice, this might be unavailable due to privacy and storage issues. To address this, \cite{chundawat2023zero} introduced zero-shot machine unlearning, which does not require access to the training dataset or knowledge about the training process. This approach builds on top of error-maximization-based noise generation~\cite{tarun2023fast}, using error-minimized noise to mimic the retain dataset and gated knowledge transfer from an expert teacher. Conversely, SOTA approaches like~\cite{chundawat2022can,graves2021amnesiac} rely on retraining or fine-tuning to enable unlearning. SSD~\cite{foster2023fast} is a recently proposed retraining-free post-hoc approach to unlearning, which uses Fisher Information Matrix (FIM)~\cite{kirkpatrick2017overcoming} to induce forgetting by dampening the parameters important to the forget set. ASSD~\cite{schoepf2024parametertuningfree} further generalizes these parameters dampening approach.

Methods like~\cite{cha2024learning, fan2024salun, jia2024model} focus on instance-level unlearning, while our method ({\pname}) focuses on class or sub-class unlearning, similar to~\cite{lin2023erm, tarun2023fast, foster2023fast, schoepf2024parametertuningfree}. {\pname} is closely related to zero-shot and retraining-free unlearning approaches; however, it requires neither the retain/forget set nor fine-tuning/retraining to achieve unlearning. {\pname} focuses on learning class/sub-class specific visual representations, enabling selective unlearning on demand without requiring any ad-hoc, expensive post-processing. To the best of our knowledge, {\pname} is the first unlearning approach that enables on-the-fly forgetting by only specifying the name of the forget class(es). This is possible because {\pname} prepares for future unlearning requests during training itself, making it an unlearning-aware approach.

\begin{figure*}[t]
    \centering
    \includegraphics[width=0.88\textwidth, height=6.2cm]{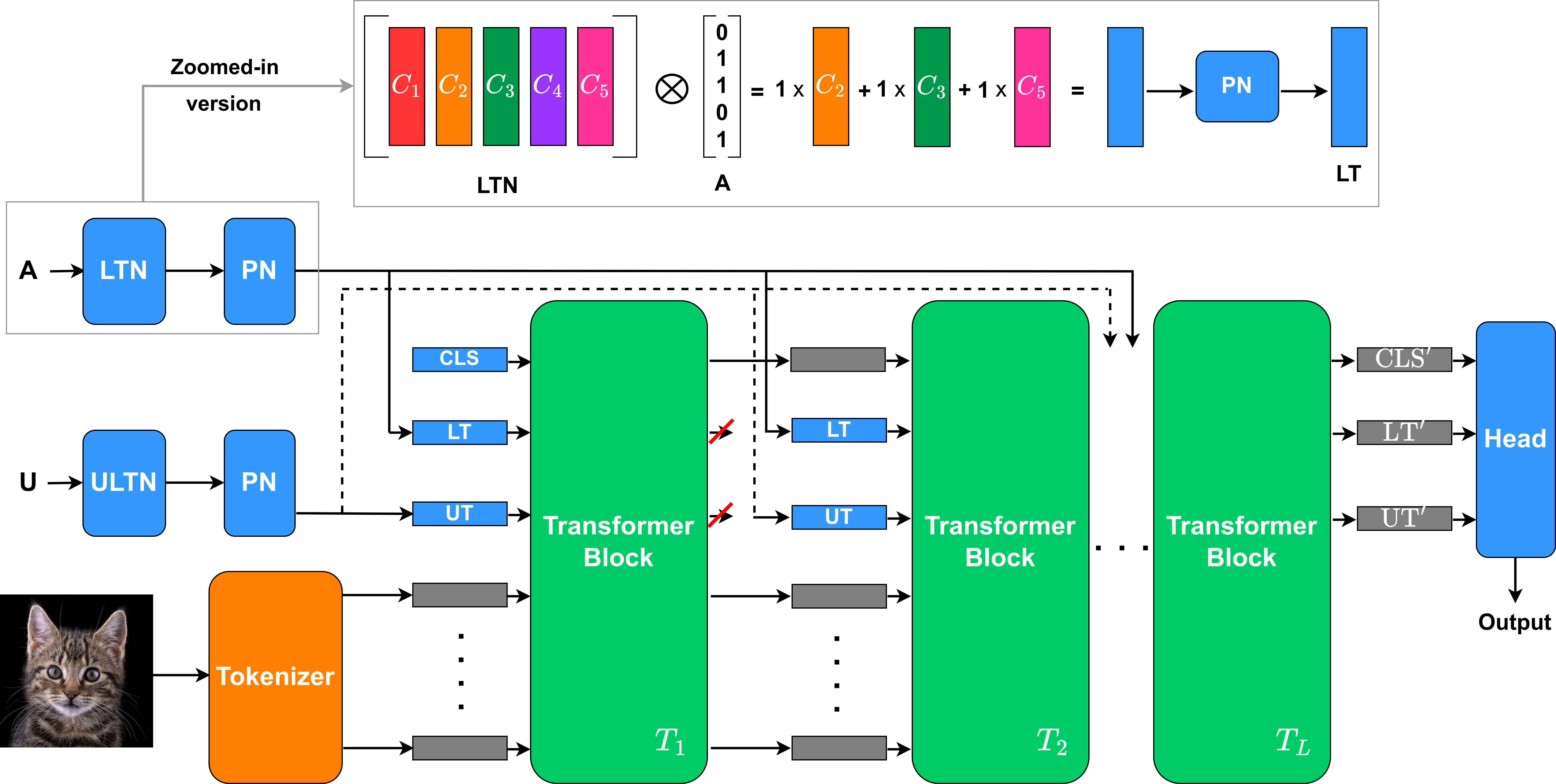}
    \caption{The overview of the proposed architecture: The model accepts multi-hot encoding $A$ for the retain classes. This is processed by $LTN$ and $PN$ (operation shown in the zoomed-in version; note that $ULTN$ and $PN$ have a similar zoomed-in version), and it produces output $LT$. The same operation is performed on the forget class set encoding $U$ and it outputs $UT$. This information is added to the input token in all the layers. The final layer outputs $CLS$, $LT'$, and $UT'$ are concatenated and used for the final classification layer. Here, $C_i$ is the \emph{learnable} class key for the $i^{th}$ class.}
    \label{fig:enter-label}
    
\end{figure*}
\section{Proposed Approach}\label{sec:proposed_approach}

We propose an unlearning-compliant network that can forget classes or sub-classes on-the-fly without retraining or fine-tuning. The entire process simply requires a class key to control the behavior for specific class(es) or sub-class(es), which are learned during training. The following sections discuss the approach in detail.

\subsection{Notations}

Consider we have a vision transformer \cite{dosovitskiy2020image,liu2021swin,DBLP:journals/corr/abs-2103-17239} based model $\mathcal{M}$ with parameters $\theta$ spread over $L$ layers, each with parameters ${\theta}_l$, where $l \in \{1,2,\dots, L\}$. Let $\cD=\{I_i,y_i\}_{i=1}^K$ denote a dataset, where $I_i \in \mathbb{R}^{M\times N \times 3}$ is an input image and $y_i \in \mathcal{Y}$ its label. The following sections detail the training step.

\subsection{Batch Class Distribution}

Let $\bx_i \in \mathbb{R}^{k\times d}$ represent the tokenized input for image $I_i$ obtained using the tokenizer of a vision transformer. Assume a batch of size $B$ contains $\mathcal{Y}_a$ unique labels (proxy \textit{retain class set}), while the remaining classes $\mathcal{Y}\setminus\mathcal{Y}_a$ (proxy \textit{forget class set}) are absent in the current batch. In the following, we use $y$ to denote both the label of image $\bx$ and its corresponding one-hot representation. To denote the unique classes present/absent in a batch using a single representation, we use the following multi-hot representations:\vspace{-1mm}
\begin{align}
\small
    \label{eq:class_dist}
    A= \sum_{y\in \mathcal{Y}_a} y \quad \& \quad
    U = \mathds{1} - A
\end{align}
where $A \in \{0,1\}^{|\mathcal{Y}|}$ and $U \in \{0,1\}^{|\mathcal{Y}|}$ are multi-hot vectors of size $|\mathcal{Y}|$, representing retain and forget class sets, respectively. As we only consider unique labels, the summation of one-hot vectors always produces a binary vector.

\subsection{Incorporating Prompt in the Input}

In the previous section, we described encoding about class present or absent in a given batch. To integrate these embeddings into the model, we transform them into learnable prompts (keys), concatenated with the input tokens. The operations are defined as follows:
\begin{align}
\small
\label{eq:learntokenet}
    & LT =PN_{\eta_1}(LTN_{\phi}(A))\\
    & UT =PN_{\eta_2}(ULTN_\xi(U_h \lor U))
\label{eq:unlearntokenet}
\end{align}
where $LT$ and $UT$ are the learnable prompts for the retain and forget class sets, respectively. $\lor$ denotes $OR$ operation, and $U_h$ is a zero-initialized frozen network parameter of the same size as $U$. $LTN$ and $ULTN$ denote the retain and forget token networks, respectively. The projection networks $PN_{\eta_1}$ and $PN_{\eta_2}$ project the token network output to a compatible dimension. In addition to the base model, $\phi$, $\xi$, $\eta_1$, and $\eta_2$ are the learnable parameters of the fully connected layers. We inject these prompts into the input token $\bx$, where $\bx[0,:]$ is the $CLS$ (class) token and $\bx[1:,:]$ contains the image tokens. After incorporating the learned and unlearned tokens, the new input becomes:
\begin{equation}
\small
    \label{eq:input_prompt}
    \bx=Concat(\bx[0,:], LT, UT, \bx[1:,:])
\end{equation}
Here, the new input tokens contain the information on the classes we want to learn and those we want to forget.

\subsection{Deep Prompting}

In Eq.~\ref{eq:input_prompt}, we concatenate the prompts to the input tokens, which is often not sufficient to incorporate the injected information into the deeper layers. Thus, the model's performance starts to degrade as we increase the number of layers in the transformer. We address this by employing the deep prompting strategy of~\cite{jia2022vpt} with a few modifications. Let $\cM_{\theta_l}$ be the $l^{th}$ layer operation with respect to the $l^{th}$ layer parameter. The operations at the first encoder layer are defined as follows:
\begin{equation}
\small
    H_0 =\cM_{\theta_0}(\bx)
\end{equation}
where $H$ is the hidden state obtained after layer $0$. Similarly, at a deeper layer $l$, the operations can be defined as:
\begin{equation}
\small
    H_{l+1} =\cM_{\theta_l}(H_l)
\end{equation}
However, $H_l$ is defined similarly to Eq.~\ref{eq:input_prompt}, i.e.,
\begin{equation}
    \label{eq:deep_prompt}
    \small
    H_l=Concat(H_l[0,:], LT, UT, H_l[3:,:])
\end{equation}
We use the same learned and unlearned prompts at all layers $l$, where $l\in \{1,2,\dots L\}$. In the ablations, we illustrate that this approach is both cost-effective and beneficial compared to disjoint parameters for each layer, which is different from the observations made in~\cite{jia2022vpt}.

In the classical ViT~\cite{dosovitskiy2020image} architecture, the hidden representation of the $CLS$ token is passed to the final classification layer. However, in our case, we also need to inject the knowledge of the retain and forget sets into the classifiers. Thus, the final classification layer $FC:\mathbb{R}^{3\times d}\rightarrow\mathbb{R}^N$ is defined as:
\begin{equation}
\small
    \label{eq:classifier}
    logits=FC(H_L[:3,:])
    \vspace{-2mm}
\end{equation}
In the ablations, we show the impact of Eq.~\ref{eq:classifier} on the final model's performance.

\subsection{Drop and Expansion Strategy}

In previous sections, we assumed that during training, the retain class set contains classes present in a batch, while the forget class set contains the remaining classes. This creates a strong correlation between the ground-truth labels and the retain/forget class sets, causing it to fail when, at test time, we include the ground truth of a sample in the forget class set. Our aim is to control class prediction based on the presence or absence of class prompts or keys. To address this, we propose the random class drop and expansion strategy, which strongly discourages the correlation between the ground-truth labels and the retain/forget class sets.

Let us assume that $\mathcal{Y}$ denotes the set of all classes present in the dataset, and $\mathcal{Y}_a$ represents the unique classes present in the current batch. Our strategy employs the following steps to generate the retain and forget class sets for that batch:
\begin{itemize}
    \item We choose a random number $r_a$ $\in$ [0, $|\mathcal{Y}_a|$) to identify the number of classes in the drop set $\cD$. Then, we randomly choose $r_a$ classes from $\mathcal{Y}_a$ and add them to $\cD$. Similarly, we choose another random number $r_u$ $\in$ [0, $|\mathcal{Y}\setminus\mathcal{Y}_a|$) to identify the number of classes in the expansion set $\cE$, randomly choose $r_u$ classes from $\mathcal{Y}\setminus\mathcal{Y}_a$, and add to $\cE$.
    \item We create the new retain class set $\mathcal{Y}_a$ as follows:
    \begin{align}
    \small
    \mathcal{Y}_a \leftarrow \mathcal{Y}_a \setminus \cD \quad \& \quad
    \mathcal{Y}_a \leftarrow \{\mathcal{Y}_a \cup \cE\}
    \vspace{-4mm}
    \end{align}
\end{itemize}
Here $\mathcal{Y}\setminus\mathcal{Y}_a$ is the forget class set. The new retain set $\mathcal{Y}_a$ includes labels from samples not present in the batch. It also excludes labels for which samples are actually present. Using Eq.~\ref{eq:class_dist}, we generate the binary vectors representing the retain ($A$) and forget ($U$) class sets. This strategy provides a key advantage during the model's training, where the model's batch optimization is not sample-dependent but rather becomes prompt-dependent. In the next section, we discuss our optimization strategy.

\subsection{Unlearning Aware Optimization}

For samples $\{\bx_i,y_i\}_{i=1}^B$ in a batch of size $B$, we use three different loss functions - Cross Entropy, Mean Squared Error (MSE), and Inverse Cross Entropy. We use the binary vectors $A$ and $U$ to represent the retain and forget class sets, respectively. Let ${I}(v)$  is an indicator function defined as:
\begin{equation}
    {I}(v)=\begin{cases} 
      1 &, v > 0 \\
      0 &, otherwise
   \end{cases}
\end{equation}

\subsubsection{Retain Set Loss}

We minimize the Cross-Entropy (CE) loss for samples whose labels are present in the retain class set. With slight abuse of notation, we use $y_i$ to denote both the label of $x_i$ and its one-hot representation. The CE loss can then be written as:
\begin{equation}
\small
    \label{eq:learn_loss}
    \cL_{CE}=\frac{1}{\sum_{i=1}^B{I(y_i \boldsymbol\cdot A)}}\sum_{i=1}^B{I}(y_i \boldsymbol\cdot A) 
    \times CE(\cM_\theta(x_i),y_i)
\end{equation}
Here $\boldsymbol\cdot$ denotes the dot product between two vectors and $y_i\in\{0,1\}^{\mathcal{Y}}$. Eq.~\ref{eq:learn_loss} enables learning for samples in a batch whose labels exist in the retain class set.

\subsubsection{Forget Set Loss} 

Let $p_i=\cM_\theta(\bx_i)$ be the logits, predicted by the model, for an input $\bx_i$. For this loss function, we first select the logits corresponding to the forget class set and then force them to follow a uniform distribution. To accomplish this, we compute the modified logits $\hat{{p}_{i}}$ as follows: \vspace{-2mm}
\begin{equation}
    \hat{{p}_{i}}=p_i\odot U
\end{equation}
Here, $\odot$ is the element-wise multiplication, and it masks the logits corresponding to the retain class set. We can now define the MSE loss as follows: \vspace{-2mm}
\begin{equation}
    \label{eq:mse}
    \cL_U=MSE\left(\hat{{p}_{i}}, \frac{1}{\sum_{i=1}^{|\cY|}U_i}\right)
\end{equation}
This loss forces the logits, corresponding to the forget class set, to follow a uniform distribution, and this can be considered equivalent to forgetting.

\subsubsection{Inverse Cross Entropy}

The MSE loss assigns equal weight to all classes in the forget class set. 
Moreover, once the model forgets a class, its samples show high scores with the next probable class(es). Therefore, we introduce another loss function that forces the sample to forget its ground-truth label class if that class is in the forget class set. We achieve this by maximizing the cross-entropy loss for that sample as:
\begin{equation}
\small
    \label{eq:complementary}
    \cL_{I}=\left({\frac{1}{\sum_{i=1}^B{I(y_i \boldsymbol\cdot U)}}\sum_{i=1}^BI(y_i \boldsymbol\cdot U) 
    \times CE(\cM_\theta(x_i),y_i)}\right)^{-1}
\end{equation}
Here $\boldsymbol\cdot$ denotes the dot product between two vectors. The loss becomes smaller as the cross-entropy loss for $y_i$ increases and this is equivalent to forgetting the sample's ground truth. We illustrate the impact of $\cL_{I}$ in the ablations. 

\begin{table}[t]
  \centering
  \fontsize{12pt}{14pt}\selectfont

    \caption{Single-class forgetting on CIFAR100 and CIFAR20 datasets with 224 $\times$ 224 resolution using ViT-B/16 architecture. The higher $\cA_{r}$ and lower $\cA_{f}$ and $MIA$ are better. $M$: Mean}
  \label{tab:fullclass}
  \vspace{-0.5em}
    \resizebox{0.47\textwidth}{!}{

\begin{tabular}{l|l|c|c|c} \hline
  
    \multicolumn{1}{l|}{Selected Classes} & Methods & $M(\mathcal{A}_{r}) \uparrow$ & $M(\mathcal{A}_{f}) \downarrow$ & $M(MIA) \downarrow$ \\ \hline
\multicolumn{1}{l|}{} & baseline & 88.9 & 93.5 & 87.8 \\
\multicolumn{1}{l|}{} & retrain & 90.1 & 0.0 & 7.2 \\
\multicolumn{1}{l|}{baby,} & finetune & 80.8 & 0.6 & 17.3 \\
\multicolumn{1}{l|}{lamp,} & bad teacher & 87.5 & 23.4 & \textbf{0.1} \\
\multicolumn{1}{l|}{mushrooms,} & UNSIR & 88.6 & 47.2 & 20.7 \\
\multicolumn{1}{l|}{rocket,} & amnesiac & 88.3 & 0.0 & 1.4 \\
\multicolumn{1}{l|}{sea} & SSD & 88.7 & 7.4 & 2.0 \\
\multicolumn{1}{l|}{} & ASSD & 88.1 & 0.0 & 2.4 \\  \cmidrule{2-5}   
\multicolumn{1}{l|}{} & \textbf{{\pname}} & \textbf{91.1} & \textbf{0.0} & 2.1 \\ \hline
\multicolumn{1}{l|}{} & baseline & 95.7 & 96.3 & 88.3\\
\multicolumn{1}{l|}{} & retrain & 94.7 & 0.0 & 8.7\\
\multicolumn{1}{l|}{electrical devices,} & finetune & 85.9 & 0.2 & 20.3\\
\multicolumn{1}{l|}{natural scenes,} & bad teacher & 93.3 & 4.6 & \textbf{0.1}\\
\multicolumn{1}{l|}{people,} & UNSIR & 93.2 & 75.3 & 43.6\\
\multicolumn{1}{l|}{vegetables,} & amnesiac & 93.5 & 0.0 & 1.1\\
\multicolumn{1}{l|}{vehicle2} & SSD & 94.7 & 10.7 & 2.7\\
\multicolumn{1}{l|}{} & ASSD & 94.9 & 0.0 & 3.6\\ \cmidrule{2-5}    
\multicolumn{1}{l|}{} & \textbf{{\pname}} & \textbf{96.0} & \textbf{0.0} & 0.2\\\hline
\end{tabular}
}
\vspace{-4mm}
\end{table}

\subsubsection{Final Objective} 

Finally, we leverage Eq.~\ref{eq:learn_loss}, \ref{eq:mse}, and \ref{eq:complementary} to create our final loss function. The joint objective is:
\begin{equation}
    \label{eq:joint}
    \cL_{\theta, \phi,\xi,\eta_1,\eta_2}= \beta\cL_{CE} + \gamma\cL_U+\tau\cL_{I}
\end{equation}
Eq.~\ref{eq:joint} is jointly optimized with respect to the parameter $\Theta=[\theta, \phi,\xi,\eta_1,\eta_2]$, where $\theta$ is the base model parameter, and $\phi,\xi,\eta_1,\eta_2$ are the additional parameters added to the model. The added parameters $\phi,\xi,\eta_i$ contain only a tiny fraction of the total number of parameters $(\approx 3\%)$ compared to the base model. The ablations over the hyperparameters $\beta,\gamma$, and $\tau$ are shown in Sec.~\ref{sec:ablations}. \emph{It is important to note that the base model parameters $(\theta)$ are optimized together with the additional parameters, ensuring that the original weights, on their own, lack predictive capability. This approach helps mitigate potential attacks, including privacy-related concerns.}

{\pname} is both zero-shot~\cite{chundawat2023zero} for the training samples and training-free~\cite{foster2023fast}. Our model is also zero-shot for the forget samples. To the best of our knowledge, this is the first work in this area that does not require any forgetting samples for unlearning (though~\cite{tarun2023fast} also does not use forget samples, they still have to learn a generator to exclusively generate synthetic forget set samples to train another model for unlearning). The inference step is simple and please refer to the supplementary material for the inference step.

\begin{table*}[t]
\small
    \scriptsize
    \setlength{\tabcolsep}{3mm}
    \centering
        \caption{The results for the CIFAR10/100 and TinyImageNet-200 dataset using ViT-B/16 architecture. Resolutions: CIFAR10/100: $224\times 224$ and TinyImageNet-200: $384\times 384$. $C_f$ denotes \#number of classes to forget. Higher $\mathcal{A}_{r}$ is better, lower $\mathcal{A}_{f}$ is better.}
    \label{tab:vitb16}
    \resizebox{\textwidth}{!}{
    
    \begin{tabular}{ c|l|l|l|| c|l|l|l|| c|l|l|l }
    \toprule
        \multicolumn{4}{c||}{\textbf{CIFAR10}} &\multicolumn{4}{c||}{\textbf{CIFAR100} }&\multicolumn{4}{c}{\textbf{TinyImageNet}} \\
        \toprule
        \shortstack{$C_f$} & Method &$\cA_r \uparrow$ & $\cA_f \downarrow$ & \shortstack{$C_f$} & Methods &$\cA_r \uparrow$ & $\cA_f \downarrow$ & \shortstack{
        $C_f$} & Methods &$\cA_r \uparrow$ & $\cA_f \downarrow$\\
        \midrule
        \multirow{4}{*}{0} & SSD   &\textbf{98.8}&  -- & \multirow{4}{*}{0}  & SSD   &90.9&  --& \multirow{4}{*}{0} & SSD   & \textbf{85.8}&  -- \\
        & ASSD   &\textbf{98.8}&  -- &   & ASSD   &90.9&  --&   & ASSD   &\textbf{85.8}&  -- \\
        & SalUn   &\textbf{98.8}&  -- &   & SalUn   &90.9&  --&  & SalUn   &\textbf{85.8}&  -- \\ \cmidrule{2-4} \cmidrule{6-8} \cmidrule{10-12}
        & \textbf{{\pname}}   &97.9&  -- &   & \textbf{{\pname}}   &\textbf{91.8}&  -- & & \textbf{{\pname}}   &84.8&  -- \\
        \midrule
        
        \multirow{4}{*}{1} & SSD   & \textbf{98.8} & 4.4 & \multirow{4}{*}{10}  & SSD   &80.2&  32.2& \multirow{4}{*}{20} & SSD   & 82.1 &  13.4 \\
        & ASSD   & 95.8 & 3.3 &   & ASSD   &83.9&  38.0&   & ASSD   & 80.7 & 5.7 \\
        & SalUn   &98.3&  0.1 &   & SalUn   &89.7&  16.4&  & SalUn   &77.6&  8.9 \\ \cmidrule{2-4} \cmidrule{6-8} \cmidrule{10-12}
        & \textbf{{\pname}}   &97.9&  \textbf{0.0} &   & \textbf{{\pname}}   &\textbf{91.8}&  \textbf{0.3} & & \textbf{{\pname}}   &\textbf{84.6}&  \textbf{0.0} \\
        \midrule
        
        \multirow{4}{*}{4} & SSD   & 99.0 &  98.5 & \multirow{4}{*}{40}  & SSD   &91.1&  90.4& \multirow{4}{*}{80} & SSD   & \textbf{84.9} &  87.1 \\
        & ASSD   & 97.5 & 97.9 &   & ASSD   &84.9&  24.9&   & ASSD   & 81.9 &  17.8 \\
        & SalUn   &\textbf{99.4}&  10.8 &   & SalUn   &88.6&  8.2&  & SalUn   &70.7&  4.6 \\ \cmidrule{2-4} \cmidrule{6-8} \cmidrule{10-12}
        & \textbf{{\pname}}   &99.2&  \textbf{0.0} &   & \textbf{{\pname}}   &\textbf{93.2}&  \textbf{0.3} & & \textbf{{\pname}}   &{84.7}&  \textbf{0.0} \\
        \midrule
        
        \multirow{4}{*}{8} & SSD   & 99.1 &  98.7 & \multirow{4}{*}{80}  & SSD   &91.0& 90.8& \multirow{4}{*}{160} & SSD   & 84.7 &  86.0 \\
        & ASSD   & 98.9 & 78.5 &   & ASSD   &93.0&  40.7&   & ASSD   & 85.7 &  24.7 \\
        & SalUn   &99.5&  8.0 &   & SalUn   &84.9&  3.9&  & SalUn   &71.1 &  4.0 \\ \cmidrule{2-4} \cmidrule{6-8} \cmidrule{10-12}
        & \textbf{{\pname}}   &\textbf{99.6}&  \textbf{0.0} &   & \textbf{{\pname}}   &\textbf{96.6}&  \textbf{0.0} & & \textbf{{\pname}}   &\textbf{88.2}&  \textbf{0.0} \\
        \midrule
    \end{tabular}
    }
\end{table*}

\begin{table*}[t]
\scriptsize
    \setlength{\tabcolsep}{3mm}
    \centering
     \caption{The results for the CIFAR20 $(224 \times 224)$ dataset using ViT-B/16 architecture. Higher $\mathcal{A}_{r}$ is better, lower $\mathcal{A}_{f}$ is better.}
    \label{tab:vitb16_subclass}
    \scriptsize
    \resizebox{\textwidth}{!}{
     \begin{tabular}{ c|l|l|l|l|l|l}
     \toprule
     Selected Sub-classes & Metric & baseline & retrain & ASSD & SSD & \textbf{NOVO}\\
     \toprule
     baby, lamp, mushrooms, & $Mean(\mathcal{A}_{r}) \uparrow$ & 95.7& 94.6 & 94.3& 95.5 & \textbf{96.7}\\
     rocket, sea & $Mean(\mathcal{A}_{f}) \downarrow$  & 95.4 &54.3 & 27 & 43.5 & \textbf{14.4}\\
     \toprule
     \end{tabular}
    }
\end{table*}


\section{Experiments and Results}\label{sec:experiments_and_results}


To demonstrate the efficacy of {\pname}, we conduct extensive experiments on medium-scale (CIFAR10/100) and large-scale (TinyImageNet-200)~\cite{Le2015TinyIV} datasets with varying image resolutions (32$\times$32, 224$\times$224, 384$\times$384) and architectures (ViT~\cite{dosovitskiy2020image}, CAIT~\cite{DBLP:journals/corr/abs-2103-17239} and SWIN~\cite{liu2021swin}). We compare {\pname} with recent transformer-based unlearning approaches like ~\cite{foster2023fast,schoepf2024parametertuningfree,fan2024salun}. The following sections cover our results on class and sub-class forgetting.

\subsection{Single-class Forgetting}

Single-class forgetting~\cite{foster2023fast,schoepf2024parametertuningfree} is the most popular use-case for unlearning methods. Similar to the experimental setup used in~\cite{foster2023fast}, we demonstrate our results using the ViT-B/16 architecture~\cite{dosovitskiy2020image} on selected classes of the CIFAR20 and CIFAR100 datasets~\cite{krizhevsky2009learning}. We compare {\pname} against the following methods - (a) $baseline$: model trained on retain ($D_r$) and forget ($D_f$) datasets, (b) $retrain$: model trained on $D_r$, (c) $finetune$: baseline model fine-tuned on $D_r$ for 5 epochs, (d) $bad\ teacher$~\cite{chundawat2022can}, (e) $amnesiac$~\cite{graves2021amnesiac}, (f) $UNSIR$~\cite{chundawat2023zero}, (g) $SSD$~\cite{foster2023fast}, and (h) $ASSD$~\cite{schoepf2024parametertuningfree}. We use the notation $\cA_r$ and $\cA_f$ to denote the accuracies on the retain and forget classes of the test set, respectively. We also investigate whether the information about the forget set is still present in the model using the $MIA$ metric. For $MIA$, we use the logistic regression implementation from~\cite{foster2023fast}.
The results for the CIFAR100 and CIFAR20 datasets using the ViT-B/16 architecture and $224\times224$ resolution are shown in Table~\ref{tab:fullclass}. In this table, we can observe that {\pname} shows consistently better results compared to the recent baselines. We also find that the proposed model shows better results on $D_r$ compared to the retrain baseline due to the prompt-based architectural modification.
For the $MIA$ metric, {\pname} is highly competitive as well. 

\begin{table*}[t]
    \centering
    \scriptsize
    \setlength{\tabcolsep}{5.8mm}
  
      \caption{Multi-class Forgetting on the CIFAR100 dataset (32 $\times$ 32 resolution) using the CAIT architecture. $C_f$ denotes the number of classes to forget. Random Multi-class Forgetting on the CIFAR100 dataset (224 $\times$ 224 resolution) using the ViT-B/16 architecture. $C_f$ denotes the number of classes to forget. Higher $\mathcal{A}_{r}$ is better, and lower $\mathcal{A}_{f}$ is better.}
    \scriptsize
     \resizebox{\textwidth}{!}{
    \begin{tabular}{c|l|l|l|c|l|l|l} \toprule
      \multicolumn{4}{c|}{\textbf{CIFAR100 (CAIT Architecture)}} &\multicolumn{4}{c}{\textbf{CIFAR100 (Random)} }\\ \toprule
      
      $C_f$ & Methods & $\mathcal{A}_{r} \uparrow$ & $\mathcal{A}_{f} \downarrow$ & $C_f$ & Methods & $\mathcal{A}_{r} \uparrow$ & $\mathcal{A}_{f} \downarrow$ \\ \toprule
      
       \multirow{4}{*}{0} & SSD & 67.1 & --&\multirow{4}{*}{0} & Retrain & -- & --\\
       &  ASSD & 67.1 & --& & SSD & 90.9  & -- \\
       & SalUn & 67.1 &--& & ASSD & 90.9  & -- \\ 
       \cmidrule{2-4}  \cmidrule{6-8}
       & \textbf{{\pname}} & \textbf{71.0}  & --& & \textbf{{\pname}} & \textbf{91.8}  & --\\\hline
      \multirow{4}{*}{10} & SSD & 66.7 & 23.0 & \multirow{4}{*}{10} &Retrain & 92.0 $\pm$ 0.3 & \textbf{0.0}\\
       & ASSD & 52.1 & 1.8 & & SSD & 85.9 $\pm$ 5.4 & 19.6 $\pm$ 7.7\\
       & SalUn & 64.8 & 16.9 & & ASSD & 81.2 $\pm$ 4.1  & 5.6 $\pm$ 4.5\\ 
       \cmidrule{2-4} \cmidrule{6-8}
       & \textbf{{\pname}} & \textbf{71.6} & \textbf{0.1} & &\textbf{{\pname}} & \textbf{92.3 $\pm$ 0.3} & 0.1 $\pm$ 0.1 \\\hline
      \multirow{4}{*}{40} & SSD & 67.2 & 66.7 &\multirow{4}{*}{20} & Retrain & 92.6 $\pm$ 0.7 & \textbf{0.0} \\
       & ASSD & 66.1 & 14.5 & &SSD & 91.2 $\pm$ 0.2 & 82.1 $\pm$ 5.6\\
       & SalUn &64.3&18.1& &ASSD & 86.72 $\pm$ 0.5 & 21.2 $\pm$ 7.1\\
       \cmidrule{2-4} \cmidrule{6-8}
       & \textbf{{\pname}} & \textbf{76.1} & \textbf{0.2} &  & \textbf{{\pname}} & \textbf{92.8 $\pm$ 0.5} & 0.3 $\pm$ 0.1\\\hline
      \multirow{4}{*}{80} & SSD & 67.8 & 66.8 & \multirow{4}{*}{50} & Retrain & \textbf{94.5 $\pm$ 0.9} & \textbf{0.0}\\
       & ASSD & 73.3 & 22.8 & & SSD & 91.8 $\pm$ 0.2 &  90.3 $\pm$ 0.1\\
       & SalUn & 66.0 &11.3& & ASSD & 86.4 $\pm$ 2.2 & 35.7 $\pm$ 4.9\\ 
       \cmidrule{2-4} \cmidrule{6-8}
       & \textbf{{\pname}} & \textbf{88.1} & \textbf{0.0}  &&\textbf{{\pname}} & 94.1 $\pm$ 1.0 & 0.2 $\pm$ 0.1\\
       \bottomrule
    \end{tabular}
    }
    \label{tab:c100random_ciat}
    \vspace{-5mm}
\end{table*}

\begin{figure}[h]
    \centering
    \includegraphics[height=4.2cm,width=8cm]{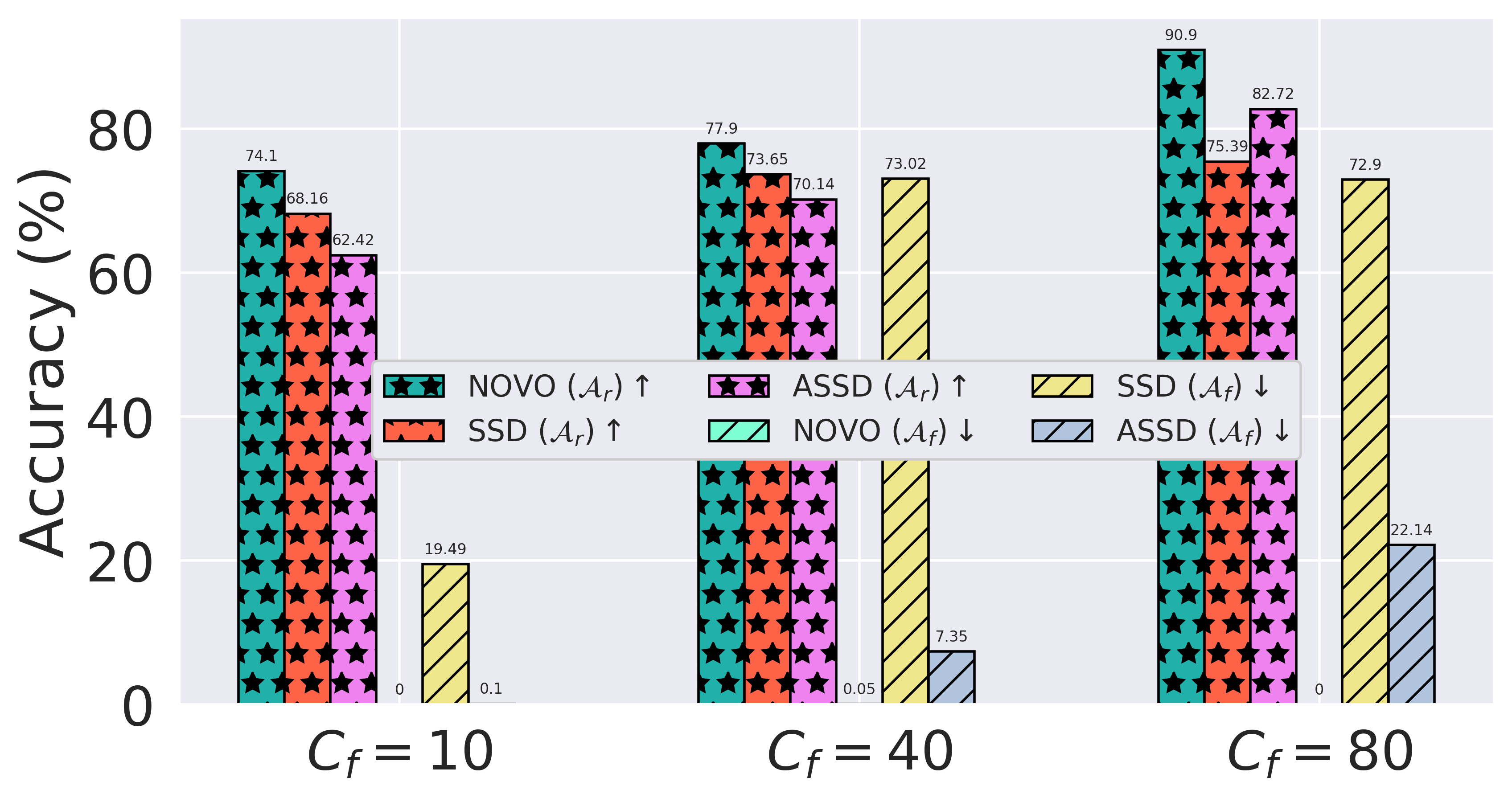}
    \caption{Multi-class Forgetting on the CIFAR100 dataset (32 $\times$ 32 resolution) using the SWIN architecture. $C_f$ denotes the number of classes to forget. Higher $\mathcal{A}_{r}$ is better, and lower $\mathcal{A}_{f}$ is better. For each $C_f$, first three bars represent $\mathcal{A}_{r}$ (in order {\pname}, SSD and ASSD), while last three represent $\mathcal{A}_{f}$.}
    \label{fig:swin_novo_ssd_assd}
\end{figure}

\subsection{Sub-class Forgetting}

Here, we drop a sub-class within a class rather than forgetting the entire class. We follow the experimental setup used in~\cite{foster2023fast} and borrow the baselines used in their paper. We illustrate our results using the ViT-B/16 architecture~\cite{dosovitskiy2020image} on selected sub-classes of the CIFAR20~\cite{krizhevsky2009learning} dataset. It should be noted that methods like~\cite{foster2023fast} implicitly assume that the sub-class label is known during unlearning, at least for the forget set. As shown in Table~\ref{tab:vitb16_subclass}, {\pname} achieves maximum forgetting, even though it is lower than the single-class case. This is because the model is still able to correctly classify a sub-class due to the presence of similar sub-classes within the class.

\subsection{Sequential Multi-class Forgetting}


We evaluate {\pname} on large-scale multi-class forgetting for CIFAR10, CIFAR100~\cite{krizhevsky2009learning} and TinyImageNet-200 datasets using the ViT-B/16~\cite{dosovitskiy2020image}, CAIT~\cite{DBLP:journals/corr/abs-2103-17239} and SWIN~\cite{liu2021swin} architectures. In this experiment, we want to forget classes with indices in [$0$, $C_f$) (hence \textit{sequential}), and we use accuracies on retain ($\cD_r$) and forget ($\cD_f$) sets to measure performance. For the CIFAR datasets, we train ViT-B/16, pre-trained on ImageNet-21k~\cite{5206848}, on images with $224\times224$ resolution. To understand whether our method works with different architectures, image resolutions, and pre-training regimes, we train CAIT, initialized using self-supervised view prediction like~\cite{Gani_2022_BMVC}, on $32\times32$ images from the CIFAR100 dataset. Similarly, we train SWIN from scratch on $32\times32$ images from CIFAR100 dataset. For TinyImageNet, we train ViT-B/16, pre-trained on ImageNet-21k, on images with $384\times384$ resolution. We compare our method with three recent SOTA baselines - SSD~\cite{foster2023fast}, ASSD~\cite{schoepf2024parametertuningfree}, and SalUn~\cite{fan2024salun}, and present the results in Table~\ref{tab:vitb16}.  As we observe in Table~\ref{tab:vitb16}, the baselines SSD~\cite{foster2023fast} and ASSD~\cite{schoepf2024parametertuningfree}, which perform well in the single-class forgetting scenario, suffer from a drastic drop in performance as $C_f$ increases. In contrast, {\pname} shows consistently better results across all datasets, and the performance does not degrade unlike the other approaches when the number of forget classes increases. The results for CIFAR100 over the CAIT~\cite{touvron2021going} architecture are shown in Table~\ref{tab:c100random_ciat} (left). Similarly, the results for CIFAR100 using SWIN~\cite{liu2021swin} architecture can be found in Fig.~\ref{fig:swin_novo_ssd_assd}. On these completely different architectures, {\pname} shows robust performance across various single/multi-class forgetting settings and outperforms the baselines by a significant margin.


\subsection{Random Multi-class Forgetting}

In the previous section, we evaluated our method by employing the experimental setup used in~\cite{lin2023erm}. Next, we want to understand how our method would perform if we randomly (and not sequentially) choose the classes to forget. We use three different seeds for randomly selecting the classes to forget. As before, we train ViT-B/16, pre-trained on ImageNet-21k~\cite{5206848}, on the CIFAR100 dataset with images of $224\times224$ resolution. We report the mean accuracies on retain ($\cD_r$) and forget ($\cD_f$) sets, along with the corresponding standard deviations, in Table~\ref{tab:c100random_ciat} (right). We observe that {\pname} generalizes to different permutations of forget-set classes.

\begin{figure}
    \centering
    \includegraphics[height=3.2cm,width=8cm]{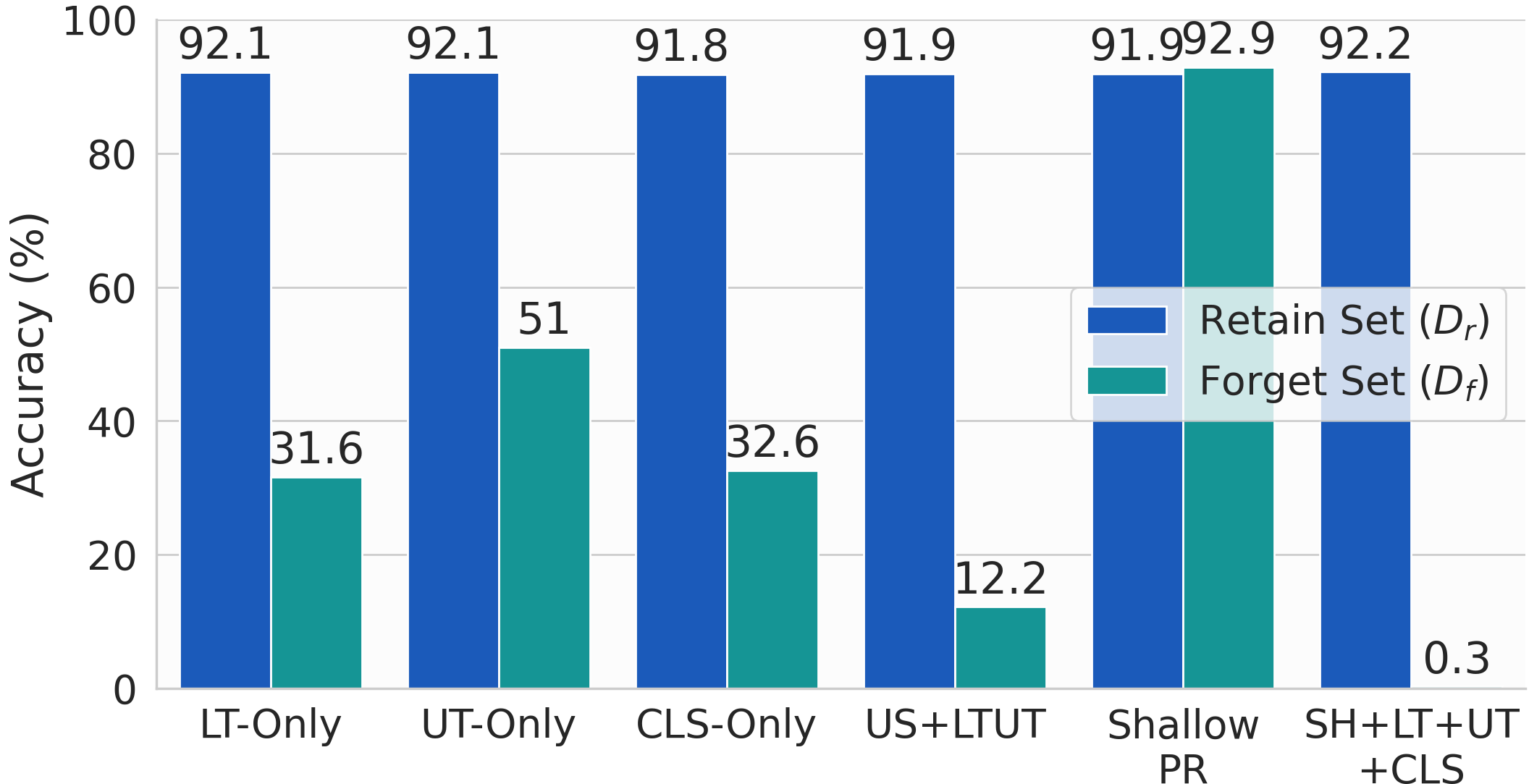}
    \qquad \qquad
    \includegraphics[height=3.2cm,width=8cm]{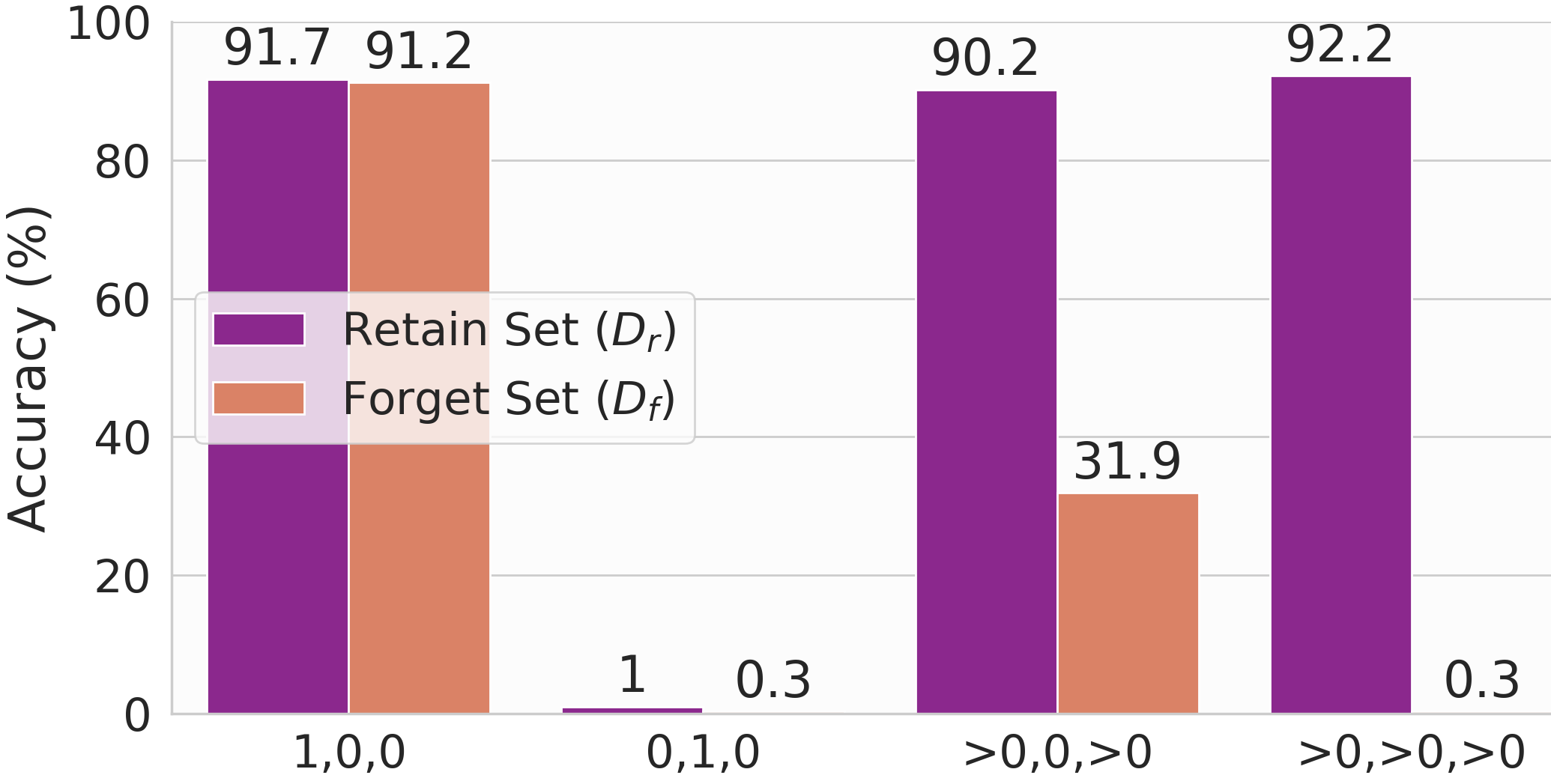}
    \caption{\textbf{Top:} Image shows the impact of prompting and token concatenation strategy for the classification. \textbf{Bottom:} Image shows the impact of hyperparameter. Here on the x-axis (a,b,c) is the value of hyperparameters ($\beta,\gamma,\tau$)}
    \label{fig:enter-label}
    \vspace{-7mm}
\end{figure}
\section{Ablations}\label{sec:ablations}

The following section illustrates the extensive ablation of the proposed components. 


\textbf{1. Drop and Expansion Strategy:} The drop and expansion strategies significantly help the model overcome sample bias since if the batch always contains the retain class samples, the model ignores the class key and predicts the output class. In Table~\ref{tab:ablations}, CIFAR100-ED, we show that the model hardly forgets anything if we exclude the drop and expansion strategies. We can observe that without drop and expansion, the accuracy on $\cD_f$ is $88.6\%$, i.e., the model does not forget at all. However, using only the drop strategy significantly improves performance and achieves accuracies of $92.1\%$ and $0.9\%$ on the retain and forget sets, respectively. Further, by adding both drop and expansion, we achieve $0.0\%$ accuracy over the forget set while preserving the retain set performance.



\textbf{2. Prompting Strategy:} In Fig.~\ref{fig:enter-label} (top), we illustrate the importance of different tokens for final classification and the impact of shallow and deep prompting strategies. More details on this can be found in the appendix.


\textbf{3. Effect of Hyperparameters $\beta$, $\gamma$, and $\tau$:} In Fig.~\ref{fig:enter-label} (bottom), we present the ablations for hyperparameters $\gamma$, $\beta$, and $\tau$. We observe that the MSE loss ($\gamma>0$) (Eq.~\ref{eq:mse}) is necessary for forgetting. Without the MSE loss, the model's accuracy on $\cD_f$ is similar to that before unlearning. The hyperparameter $\tau$ shows an interesting characteristic. We know that in a dataset with $C$ classes, we can have $2^C$ possible combination of retain/forget classes. As it is not possible to cover all scenarios with limited epochs, the model must generalize to unseen combinations of retain and forget class sets. In addition to reducing the forget set accuracy (Table~\ref{tab:ablations}), $\tau>0$ helps achieve faster convergence. The model converges in 73 epochs on the CIFAR100 dataset for $\tau>0$; however, it requires 150 epochs for $\tau=0$. Therefore, the proposed loss component plays a significant role in achieving model generalization on unseen class prompts. Table \ref{tab:ablations} also explores the importance of various components over the TinyImageNet dataset. 

\begin{table}[t]
  \centering
    \caption{Ablation studies involving forgetting 10 and 20 classes using ViT-B/16 architecture over CIFAR100 and TinyImageNet datasets. Higher $\mathcal{A}_{r}$ is better, lower $\mathcal{A}_{f}$ is better. M: Mean}
  \label{tab:ablations}
  \small
  \centering
  \resizebox{0.47\textwidth}{!}{
  \begin{tabular}{p{1.8cm}|p{2.2cm}|c|c} \toprule
    Dataset & Scenarios & $M(\mathcal{A}_{r})\uparrow$ & $M(\mathcal{A}_{f})\downarrow$ \\ \toprule
     \multirow{2}{*}{TinyImageNet} & $\gamma > 0$, $\beta > 0$, $\tau > 0$ & 84.7 & 0.0 \\
     \cmidrule{2-4}
     & $\gamma > 0$, $\beta > 0$, $\tau = 0$ in Eq.~\ref{eq:joint} & 84.7 & 21.5\\ \hline
    \multirow{2}{*}{CIFAR100-ED} & No drop or expansion strategy & 91.8  & 88.6\\
    \cmidrule{2-4}
     & Only drop strategy & 92.1 & 0.9 \\
	 \bottomrule
  \end{tabular}
  }
  \vspace{-4mm}
\end{table}

\textbf{4. Dropping Classifiers/Masking (a simple baseline): } 
A question that can arise is whether the impact of {\pname} can be achieved by just dropping the classifier weights corresponding to the forget class set. Obviously, this strategy would not work for forgetting of sub-classes as different sub-classes can be mapped to a single class. For example, in Table~\ref{tab:vitb16_subclass}, discarding classifier weights results in a drop of accuracy, on the retain dataset, from 96.7 to 76.7. Moreover, masking logits cannot actually \textit{unlearn} any information from the model; it just discards the predictions, and hence privacy concerns remain. Therefore, an attacker can easily identify the forget class set using MIA~\cite{foster2023fast}. However, our proposed method jointly trains model parameters along with the added class/sub-class keys. To predict a sample, the presence of both these parameters is compulsory; therefore, discarding the keys is equivalent to removing class/sub-class information from the model. For example, we notice that while NOVO has a mean MIA of 2.1 on CIFAR100, dropping the forget class weights from a vanilla fine-tuned model results in a mean MIA of 97 (see Table~\ref{tab:fullclass}). This is because, even after classifier deletion, the entropy characteristics for the forget set are more similar to the training set than the test set. In the appendix, we also show that, unlike logit masking, {\pname} can also be used for feature extraction.
\section{Conclusion}\label{sec:conclusion}

We propose a novel transformer-based unlearning-compliant architecture that can forget any set of classes/sub-classes on-the-fly, without any fine-tuning or training data. This approach addresses privacy concerns and is highly practical. The method associates a learnable key with each class/sub-class, and the model can predict the sample classes/sub-classes only if the corresponding key is present. Simply withdrawing the key is equivalent to forgetting that class/sub-class. During training, the model is optimized to minimize the classification loss when the key is present; otherwise, the model is trained to predict a uniform distribution. The learnable keys are defined using a deep prompting strategy, which exhibits robust performance. The proposed drop and expansion methodology helps in overcoming sample bias. We conduct extensive experiments on standard medium and large-scale datasets for ViT, CAIT, and SWIN architectures. Ablations over various components disentangle the contributions of each module. In the future, we plan to extend our method to instance-level unlearning, where we will forget individual examples instead of semantically similar clusters.

{
    \small
    \bibliographystyle{ieeenat_fullname}
    \bibliography{main}
}

\newpage
\appendix
\onecolumn




\section{What are forget classes misclassified as?}

\begin{table}[h]
    \centering
    \caption{Confusion matrix}
    \setlength{\tabcolsep}{2mm}
    \begin{tabular}{c|c|c|c|c}
         \toprule
         & 0 (aq mml) & 8 (crnvr) & 14 (ppl) & 19 (vh 2) \\\hline
         0 & & & & \\\hline
         1 (fish) & \checkmark & & & \\\hline
         2 & & & & \\\hline
         3 & & & & \\\hline
         4 & & & & \\\hline
         5 & & & & \\\hline
         6 & & & & \\\hline
         7 & & & & \\\hline
         8 & & & & \\\hline
         9 & & & & \\\hline
         10 & & & & \\\hline
         11 (hrbvr) & & \checkmark & & \\\hline
         12 & & & & \\\hline
         13 (invrtbrt) & & & \checkmark & \\\hline
         14 & & & & \\\hline
         15 & & & & \\\hline
         16 & & & & \\\hline
         17 & & & & \\\hline
         18 (vh 1) & & & & \checkmark\\\hline
         19 & & & & \\\hline
         \bottomrule    
    \end{tabular}
    \label{tab:cc}
\end{table}

Till now, we have shown that our method is able to unlearn classes, which are present in the forget class set. Now our questions is: \textit{Given an image, whose ground truth is in the forget class set, what is the class returned by our model?} Of course, this class will be in the retain class set. However, if the image of a $cat$ is classified as a $train$ consistently, it might raise suspicions and the forget dataset might get leaked unintentionally.

Our method implicitly tries to choose a class in the vicinity of the forget class set. To demonstrate this, we perform a single-class forgetting operation on $4$ superclasses of CIFAR20 - $3$ of them have similar classes in the retain class set, while $1$ does not. As is evident from Table~\ref{tab:cc}, our method mostly confuses classes \textit{aquatic mammals}, \textit{large carnivores} and \textit{vehicles 2} (labels: 0, 8 and 19) with \textit{fish}, \textit{large omnivores and herbivores} and \textit{vehicles 1} (labels: 1, 11 and 18) respectively. However, class \textit{people} (label: 14) gets confused with \textit{non-insect invertebrates} (labels: 13) as there wasn't any similar class in the retain class set.

\section{{\pname} as feature extractor}
\begin{figure}[h!]
    \centering
    \includegraphics[scale=0.6]{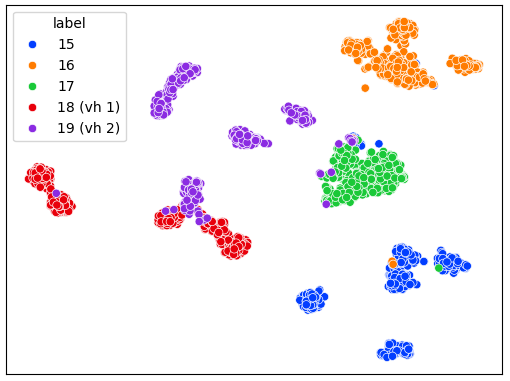}\qquad
    \includegraphics[scale=0.6]{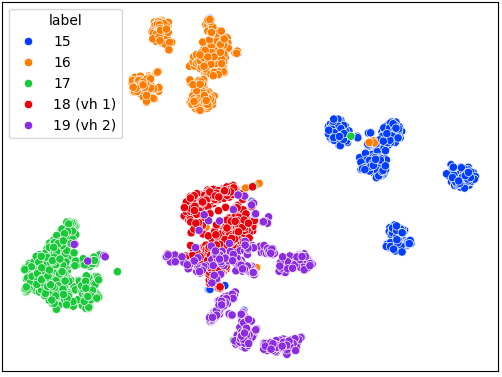}
    \caption{\textbf{Left:} Before unlearning of \textit{vehicles 1} (label: 18). \textbf{Right:} After unlearning of \textit{vehicles 1} (label: 18). }
    \label{fig:tsne}
\end{figure}
Black-box classification models, often, expose functionalities which allow feature extraction from a pre-defined intermediate layer of the model network. Moreover, the extracted features can also reveal how {\pname} unlearns classes. To demonstrate this, we use ViT-B/16~\cite{dosovitskiy2020image} architecture on the CIFAR20~\cite{krizhevsky2009learning} dataset and unlearn the \textit{vehicles 1} (label: 18) class. For a given image, we can use any token (or a combination of tokens) from the ViT as the image feature. In Fig.~\ref{fig:tsne}, we use the forget token feature to create t-SNE~\cite{JMLR:v9:vandermaaten08a} plots of classes 15 to 19 in the CIFAR20 test dataset, before and after the unlearning operation. It should be noted that \textit{vehicles 2} (label: 19) is the nearest class to \textit{vehicles 1} (label: 18).

\subsection{Before Unlearning: } In Fig.~\ref{fig:tsne} (Top), we notice that there are multiple clusters for each label in the t-SNE plot. Moreover, labels \textit{vehicles 1} (label: 18) and \textit{vehicles 2} (label: 19) have several distinct clusters and one fused cluster (as these classes are similar to one another). If we employ logit masking to perform class forgetting, the extracted features will show no change before and after the unlearning operation. Thus, despite being discriminative in the feature space, the forget class will have $0$ accuracy and an attacker may be able to guess the forget class set. Thus models, which utilize logit masking, cannot be used for feature extraction.
\subsection{After Unlearning: } In Fig.~\ref{fig:tsne} (Bottom), we notice that clusters for labels, other than \textit{vehicles 1} (label: 18) and \textit{vehicles 2} (label: 19), undergo limited changes. However, clusters of \textit{vehicles 1} (label: 18) and \textit{vehicles 2} (label: 19) fuse together which indicates that \textit{vehicles 1} class loses it's discriminative nature, in the feature space, with respect to \textit{vehicles 2}. This is also corroborated by the observation that after unlearning, \textit{vehicles 1} gets confused with \textit{vehicles 2} class around 82\% times. Thus, unlike the logit masking approach, {\pname} causes fundamental changes in the image representation and can, thus, be utilized for feature extraction.

\section{Can we freeze the base network?}
\label{base_freeze_appendix}
In traditional machine unlearning, we have access to an existing model, such as a vision transformer trained on face recognition dataset, and the objective is to fine-tune this model to unlearn a set of faces. However, in this paper, we propose an architecture, which is trained from `scratch' (from downstream task/dataset perspective) by simulating various unlearning scenarios in the given downstream task/dataset. For completeness of this paper, we demonstrate in Table~\ref{tab:freeze_base} that we can also train the prompt parameters ($\phi,\xi,\eta_1,\eta_2$), while keeping the base network $\theta$, pre-trained on the downstream task/dataset, frozen (obviously, we alter the classifier layer and train it).

\begin{figure}[h!]
    \centering
    \includegraphics[scale=0.5]{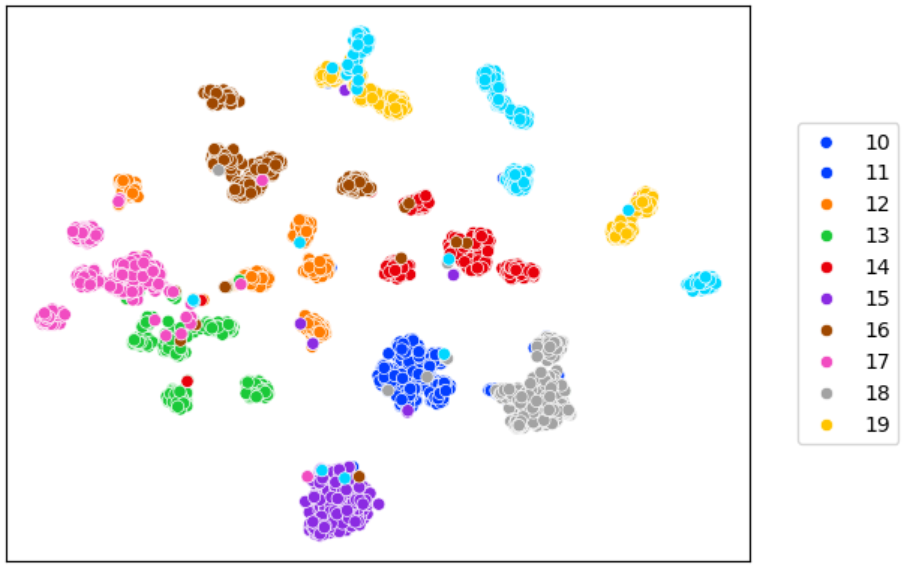}
    \includegraphics[scale=0.5]{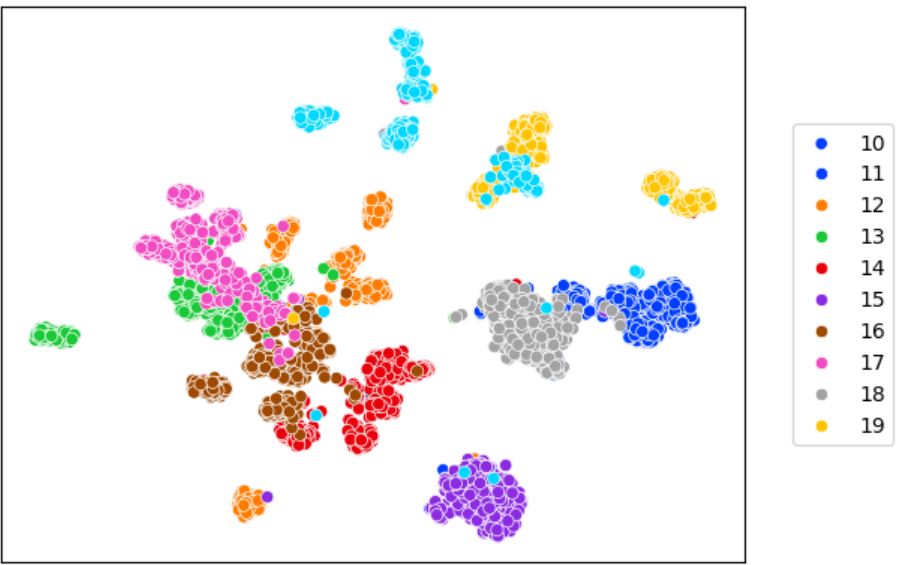}
    \caption{\textbf{Left:} Feature extracted from vanilla ViT-B/16. \textbf{Right:} Feature extracted from \pname~(minus prompts).}
    \label{fig:tsne2}
\end{figure}

However, if we choose not to train the base network $\theta$, an exploiter can use $\theta$ to extract discriminative features for unlearned classes. To demonstrate this, we use ViT-B/16~\cite{dosovitskiy2020image} architecture on the CIFAR20~\cite{krizhevsky2009learning} dataset and extract class features from two kinds of base networks - a vanilla ViT-B/16 and \pname~(without prompts). As is clear from Fig.~\ref{fig:tsne2}, vanilla ViT-B/16 features result in distinct clusters for all classes, while \pname~(without prompts) features from different classes tend to get lumped together. This shows the importance of training the base network, along with the prompt parameters.

\begin{table*}[t]
\scriptsize
    \setlength{\tabcolsep}{3mm}
    \centering
    \caption{The results on the CIFAR100 $(32 \times 32)$ dataset by training prompt parameters and classifier layer in ViT-B/16 model, pre-trained on CIFAR-100. $C_f$ denotes the \#number of classes to forget. Higher $\mathcal{A}_{r}$ is better, lower $\mathcal{A}_{f}$ is better.}
       \label{tab:freeze_base}
\resizebox{0.3\textwidth}{!}{
    \begin{tabular}{c|l|l} \toprule
        
      $C_f$ & $\mathcal{A}_{r} \uparrow$ & $\mathcal{A}_{f} \downarrow$ \\ \toprule
      
       10 & 74 & 1.2  \\
       40 & 78.7 & 3.9  \\
       80 & 88 & 3.8 \\
       \hline
       \end{tabular}}
       
    \end{table*}
\section{Prompting Strategy}

\begin{figure}
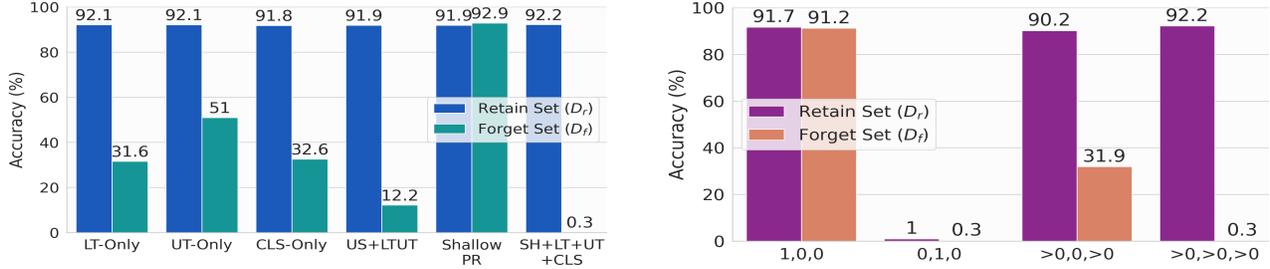

    \centering
    \includegraphics[height=3.5cm,width=8cm]{c100PR.png}
    \qquad
    \includegraphics[height=3.5cm,width=8cm]{c100HP.png}
    \caption{\textbf{Left:} Image shows the impact of prompting and token concatenation strategy for the classification. \textbf{Right:} Image shows the impact of hyperparameter. Here on the x-axis (a,b,c) is the value of hyperparameters ($\beta,\gamma,\tau$)}
    \label{fig:enter-label2}
\end{figure}

In Fig.~\ref{fig:enter-label2}, we have shown the importance of different tokens for final classification and the impact of shallow and deep prompting strategies. We observe that both $LT$ and $UT$ tokens are necessary for better performance on the forget set $\cD_f$. Using only the $LT$, $UT$, or $CLS$ tokens does not degrade the performance on $\cD_r$, but the model is unable to forget the $\cD_f$ set and shows a high accuracy of $\sim 30-50\%$. We also observe that shallow prompting ($Shallow\ PR$) is unable to forget the requested classes, and the proposed deep prompting strategy significantly helps to overcome this problem. Moreover, using $SH+LT+UT+CLS$, i.e., the $CLS$ token along with $LT$, $UT$, and deep shared prompts, the model shows complete forgetting. The impact of shared and unshared ($US+LTUT$) parameters for the token networks $LTN$ and $ULTN$ is also explored. Sharing the parameters $\phi$ and $\xi$ across all layers consistently outperforms the unshared parameter model, and the model has significantly fewer learnable parameters.

\section{Class-wise performance in Single-class Forgetting}
In Tables~\ref{tab:cw_sc_forget}-\ref{tab:cw_sc_forget2}, we present class-wise results in single-class forgetting scenario. These results were aggregated and presented in the main paper.

\begin{table*}[h]
    \centering
    \tiny
    \setlength{\tabcolsep}{3.2mm}
      \caption{Single-class forgetting on CIFAR100 with 224 $\times$ 224 resolution using ViT-B/16 architecture. The higher $\cA_{r}$ is better and lower $\cA_{f}$ and $MIA$ are better.}
    \label{tab:cw_sc_forget}



    \begin{adjustbox}{totalheight=\textheight-7\baselineskip}
    
    \begin{tabular}{l|l|l|l|l} 

        \toprule

        \multicolumn{1}{l|}{Class} & Methods & $\mathcal{A}_{r} \uparrow$ & $\mathcal{A}_{f} \downarrow$ & $MIA \downarrow$\\ 

        \midrule

        \multicolumn{1}{l|}{} & retrain~\cite{foster2023fast} & 90.27 & \textbf{0} & 21.53\\
        \multicolumn{1}{l|}{} & finetune~\cite{foster2023fast} & 80.74 & \textbf{0} & 26.77\\
        \multicolumn{1}{l|}{} & teacher~\cite{chundawat2022can} & 87.48 & 23.8 & \textbf{0}\\
        \multicolumn{1}{l|}{baby} & UNSIR~\cite{chundawat2023zero} & 88.78 & 1.97 & 14.3\\
        \multicolumn{1}{l|}{} & amnesiac~\cite{graves2021amnesiac} & 88.43 & \textbf{0} & 1.83\\
        \multicolumn{1}{l|}{} & SSD~\cite{foster2023fast} & 88.59 & \textbf{0} & 0.6\\
        \multicolumn{1}{l|}{} & ASSD~\cite{schoepf2024parametertuningfree} & 88.31 & \textbf{0} & 2.4\\
        \multicolumn{1}{l|}{} & \textbf{NOVO} & \textbf{91.2} & \textbf{0} & 2.4\\ 
        
        \midrule
        
        \multicolumn{1}{l|}{} & retrain~\cite{foster2023fast} & 90.1 & \textbf{0} & 2.27\\
        \multicolumn{1}{l|}{} & finetune~\cite{foster2023fast} & 80.25 & 0.36 & 11.77\\ 
        \multicolumn{1}{l|}{} & teacher~\cite{chundawat2022can} & 87.5 & 25.25 & \textbf{0.13}\\
        \multicolumn{1}{l|}{lamp} & UNSIR~\cite{chundawat2023zero} & 88.51 & 70.86 & 29.4\\
        \multicolumn{1}{l|}{} & amnesiac~\cite{graves2021amnesiac} & 88.43 & \textbf{0} & 2.7\\
        \multicolumn{1}{l|}{} & SSD~\cite{foster2023fast} & 89.06 & 36.89 & 0.4\\ 
        \multicolumn{1}{l|}{} & ASSD~\cite{schoepf2024parametertuningfree} & 88.64 & \textbf{0} & 1.4\\
        \multicolumn{1}{l|}{} & \textbf{NOVO} & \textbf{91.1} & \textbf{0} & 3\\ 

        \midrule
        
        \multicolumn{1}{l|}{} & retrain~\cite{foster2023fast} & 90.02 & \textbf{0} & 0.7\\
        \multicolumn{1}{l|}{} & finetune~\cite{foster2023fast} & 81.14 & 2.33 & 7.1\\ 
        \multicolumn{1}{l|}{} & teacher~\cite{chundawat2022can} & 87.42 & 12.82 & \textbf{0.03}\\ 
        \multicolumn{1}{l|}{mushroom} & UNSIR~\cite{chundawat2023zero} & 88.44 & 83.94 & 21.33\\ 
        \multicolumn{1}{l|}{} & amnesiac~\cite{graves2021amnesiac} & 88.34 & \textbf{0} & 0.47\\
        \multicolumn{1}{l|}{} & SSD~\cite{foster2023fast} & 88.82 & \textbf{0} & 3.8\\ 
        \multicolumn{1}{l|}{} & ASSD~\cite{schoepf2024parametertuningfree} & 88.07 & \textbf{0} & 2.2\\
        \multicolumn{1}{l|}{} & \textbf{NOVO} & \textbf{91.2} & \textbf{0} & 0.8\\ 

        \midrule
        
        \multicolumn{1}{l|}{} & retrain~\cite{foster2023fast} & 90.07 & \textbf{0} & 3.23\\
        \multicolumn{1}{l|}{} & finetune~\cite{foster2023fast} & 80.82 & 0.46 & 19\\
        \multicolumn{1}{l|}{} & teacher~\cite{chundawat2022can} & 87.46 & 4.2 & \textbf{0.03}\\ 
        \multicolumn{1}{l|}{rocket} & UNSIR~\cite{chundawat2023zero} & 88.47 & 65.32 & 29.13\\ 
        \multicolumn{1}{l|}{} & amnesiac~\cite{graves2021amnesiac} & 87.92 & \textbf{0} & 1\\ 
        \multicolumn{1}{l|}{} & SSD~\cite{foster2023fast} & 88.9 & \textbf{0} & 1.8\\ 
        \multicolumn{1}{l|}{} & ASSD~\cite{schoepf2024parametertuningfree} & 88.39 & \textbf{0} & 1.4\\
        \multicolumn{1}{l|}{} & \textbf{NOVO} & \textbf{91.1} & \textbf{0} & 0.8\\ 

        \midrule
        
        \multicolumn{1}{l|}{} & retrain~\cite{foster2023fast} & 90.27 & \textbf{0} & 8.43\\
        \multicolumn{1}{l|}{} & finetune~\cite{foster2023fast} & 80.82 & \textbf{0} & 21.97\\
        \multicolumn{1}{l|}{} & teacher~\cite{chundawat2022can} & 87.72 & 51.13 & \textbf{0}\\
        \multicolumn{1}{l|}{sea} & UNSIR~\cite{chundawat2023zero} & 88.85 & 13.95 & 9.1\\
        \multicolumn{1}{l|}{} & amnesiac~\cite{graves2021amnesiac} & 88.25 & \textbf{0} & 0.77\\
        \multicolumn{1}{l|}{} & SSD~\cite{foster2023fast} & 87.95 & \textbf{0} & 3.2\\ 
        \multicolumn{1}{l|}{} & ASSD~\cite{schoepf2024parametertuningfree} & 86.98 & \textbf{0} & 4.4\\
        \multicolumn{1}{l|}{} & \textbf{NOVO} & \textbf{91.1} & \textbf{0} & 3.6 \\ 

        \bottomrule

    \end{tabular}


    \end{adjustbox}

\end{table*}

\begin{table*}[h]
    \centering

    \tiny

    \setlength{\tabcolsep}{3.mm}

    \caption{Single-class forgetting on CIFAR20 with 224 $\times$ 224 resolution using ViT-B/16 architecture. The higher $\cA_{r}$ is better and lower $\cA_{f}$ and $MIA$ are better.}
    \label{tab:cw_sc_forget2}

    
    
    \begin{adjustbox}{totalheight=\textheight-7\baselineskip}

    \begin{tabular}{l|l|l|l|l} 

        \toprule

        \multicolumn{1}{l|}{Class} & Methods & $\mathcal{A}_{r} \uparrow$
        & $\mathcal{A}_{f} \downarrow$ & $MIA \downarrow$\\ 

        \midrule

        \multicolumn{1}{l|}{} & retrain~\cite{foster2023fast} & 94.71 & \textbf{0} & 9.82\\ 
        \multicolumn{1}{l|}{} & finetune~\cite{foster2023fast} & 87.35 & 0.28 & 23.6\\ 
        \multicolumn{1}{l|}{} & teacher~\cite{chundawat2022can} & 93.42 & 4.14 & \textbf{0.02}\\ 
        \multicolumn{1}{l|}{electrical} & UNSIR~\cite{chundawat2023zero} & 93.25 & 73.67 & 38.18\\ 
        \multicolumn{1}{l|}{devices} & amnesiac~\cite{graves2021amnesiac} & 93.45 & 0.03 & 1.7\\ 
        \multicolumn{1}{l|}{} & SSD~\cite{foster2023fast} & 95.82 & 53.53 & 1.32\\ 
        \multicolumn{1}{l|}{} & ASSD~\cite{schoepf2024parametertuningfree} & 95.4 & \textbf{0} & 5.72\\ 
        \multicolumn{1}{l|}{} & \textbf{NOVO} & \textbf{95.9} & \textbf{0} & 0.2\\ 

        \midrule
        
        \multicolumn{1}{l|}{} & retrain~\cite{foster2023fast} & 94.79 & \textbf{0} & 4.7\\ 
        \multicolumn{1}{l|}{} & finetune~\cite{foster2023fast} & 87.45 & 0.05 & 16.97\\ 
        \multicolumn{1}{l|}{} & teacher~\cite{chundawat2022can} & 93.5 & 2.63 & \textbf{0.06}\\ 
        \multicolumn{1}{l|}{natural} & UNSIR~\cite{chundawat2023zero} & 93.09 & 50.65 & 8.58\\ 
        \multicolumn{1}{l|}{scenes} & amnesiac~\cite{graves2021amnesiac} & 93.68 & \textbf{0} & 1.04\\ 
        \multicolumn{1}{l|}{} & SSD~\cite{foster2023fast} & 93.63 & \textbf{0} & 1.88\\ 
        \multicolumn{1}{l|}{} & ASSD~\cite{schoepf2024parametertuningfree} & 95.3 & \textbf{0} & 2\\ 
        \multicolumn{1}{l|}{} & \textbf{NOVO} & \textbf{96.2} & \textbf{0} & 0.3\\ 

        \midrule
        
        \multicolumn{1}{l|}{} & retrain~\cite{foster2023fast} & 94.54 & \textbf{0} & 1.56\\ 
        \multicolumn{1}{l|}{} & finetune~\cite{foster2023fast} & 80.18 & 0.09 & 8.08\\ 
        \multicolumn{1}{l|}{} & teacher~\cite{chundawat2022can} & 93.19 & 2.91 & \textbf{0.01}\\ 
        \multicolumn{1}{l|}{people} & UNSIR~\cite{chundawat2023zero} & 92.74 & 93.09 & 63.42\\ 
        \multicolumn{1}{l|}{} & amnesiac~\cite{graves2021amnesiac} & 93.28 & \textbf{0} & 0.6\\ 
        \multicolumn{1}{l|}{} & SSD~\cite{foster2023fast} & 95.33 & \textbf{0} & 1.2\\ 
        \multicolumn{1}{l|}{} & ASSD~\cite{schoepf2024parametertuningfree} & 95.11 & \textbf{0} & 1.48\\ 
        \multicolumn{1}{l|}{} & \textbf{NOVO} & \textbf{95.8} & \textbf{0} & 0.1\\ 

        \midrule
        
        \multicolumn{1}{l|}{} & retrain~\cite{foster2023fast} & 94.54 & \textbf{0} & 4.41\\ 
        \multicolumn{1}{l|}{} & finetune~\cite{foster2023fast} & 87.09 & 0.3 & 14.72\\ 
        \multicolumn{1}{l|}{} & teacher~\cite{chundawat2022can} & 92.92 & 8.28 & \textbf{0.02}\\ 
        \multicolumn{1}{l|}{vegetables} & UNSIR~\cite{chundawat2023zero} & 93.25 & 89.02 & 58.67\\ 
        \multicolumn{1}{l|}{} & amnesiac~\cite{graves2021amnesiac} & 93.29 & 0.02 & 1.02\\ 
        \multicolumn{1}{l|}{} & SSD~\cite{foster2023fast} & 95.71 & \textbf{0} & 1.88\\ 
        \multicolumn{1}{l|}{} & ASSD~\cite{schoepf2024parametertuningfree} & 95.07 & \textbf{0} & 2.96\\ 
        \multicolumn{1}{l|}{} & \textbf{NOVO} & \textbf{95.8} & \textbf{0} & 0.2\\ 

        \midrule
        
        \multicolumn{1}{l|}{} & retrain~\cite{foster2023fast} & 94.85 & \textbf{0} & 22.96\\ 
        \multicolumn{1}{l|}{} & finetune~\cite{foster2023fast} & 87.75 & 0.04 & 38.15\\ 
        \multicolumn{1}{l|}{} & teacher~\cite{chundawat2022can} & 93.59 & 4.88 & \textbf{0.02}\\ 
        \multicolumn{1}{l|}{vehicle2} & UNSIR~\cite{chundawat2023zero} & 93.56 & 70.31 & 48.98\\ 
        \multicolumn{1}{l|}{} & amnesiac~\cite{graves2021amnesiac} & 93.88 & \textbf{0} & 1.2\\ 
        \multicolumn{1}{l|}{} & SSD~\cite{foster2023fast} & 93.12 & \textbf{0} & 7.04\\ 
        \multicolumn{1}{l|}{} & ASSD~\cite{schoepf2024parametertuningfree} & 93.93 & \textbf{0} & 6.04\\ 
        \multicolumn{1}{l|}{} & \textbf{NOVO} & \textbf{96.3} & \textbf{0} & 0.4\\ 

        \bottomrule

    \end{tabular}
  

    \end{adjustbox}

\end{table*}

\end{document}